\documentclass[conference]{IEEEtran}
\IEEEoverridecommandlockouts
\usepackage{cite}
\usepackage{amsmath,amssymb,amsfonts}
\usepackage{algorithmic}
\usepackage{graphicx}
\usepackage{textcomp}
\usepackage{hyperref}
\usepackage{scrextend}
\usepackage{soul}

\usepackage{xcolor}
\def\BibTeX{{\rm B\kern-.05em{\sc i\kern-.025em b}\kern-.08em
    T\kern-.1667em\lower.7ex\hbox{E}\kern-.125emX}}
\begin{document}

\title{Retrieval Augmentation for Deep Neural Networks
}

\author{\IEEEauthorblockN{Rita Parada Ramos}
\IEEEauthorblockA{\textit{INESC-ID} \\
\textit{Instituto Superior Técnico}\\
\textit{Universidade de Lisboa}\\
Lisbon, Portugal \\
{\tt \scriptsize ~~~ritaparadaramos@tecnico.ulisboa.pt~~~}}

\and

\IEEEauthorblockN{Patrícia Pereira}
\IEEEauthorblockA{\textit{INESC-ID} \\
\textit{Instituto Superior Técnico}\\
\textit{Universidade de Lisboa}\\
Lisbon, Portugal \\
{\tt \scriptsize ~~patriciaspereira@tecnico.ulisboa.pt~~}}

\and

\IEEEauthorblockN{Helena Moniz}
\IEEEauthorblockA{\textit{INESC-ID} \\
\textit{Faculdade de Letras}\\
\textit{Universidade de Lisboa}\\
Lisbon, Portugal \\
{\tt \scriptsize ~~~~~~~~helena.moniz@inesc-id.pt~~~~~~~~}}

\and

\IEEEauthorblockN{Joao Paulo Carvalho}
\IEEEauthorblockA{\textit{INESC-ID} \\
\textit{Instituto Superior Técnico}\\
\textit{Universidade de Lisboa}\\
Lisbon, Portugal \\
{\tt \scriptsize ~~~~~~~~~~~~~~~~~joao.carvalho@inesc-id.pt~~~~~~~~~~~~~~~~~~}}

\and

\IEEEauthorblockN{Bruno Martins}
\IEEEauthorblockA{\textit{INESC-ID} \\
\textit{Instituto Superior Técnico}\\
\textit{Universidade de Lisboa}\\
Lisbon, Portugal \\
{\tt \scriptsize ~~~~~~bruno.g.martins@tecnico.ulisboa.pt~~~~~~}}

\and

}

\maketitle

\begin{abstract}
Deep neural networks have achieved state-of-the-art results in various vision and/or language tasks. Despite the use of large training datasets, most models are trained by iterating over single input-output pairs, discarding the remaining examples for the current prediction. In this work, we actively exploit the training data, using the information from nearest training examples to aid the prediction both during training and testing.
Specifically, our approach uses the target of the most similar training example to initialize the memory state of an LSTM model, or to guide attention mechanisms. We apply this approach to image captioning and sentiment analysis, respectively through image and text retrieval. Results confirm the effectiveness of the proposed approach for the two tasks, on the widely used Flickr8 and IMDB datasets. Our code is publicly available\footnote{\url{http://github.com/RitaRamo/retrieval-augmentation-nn}}.
\end{abstract}


\section{Introduction}

The most common methodology in deep learning involves the supervised 
training of a neural network with input-output pairs, so as to minimize a given loss function \cite{lecun2015deep}. 
In general, deep neural networks predict the output conditioned solely on the current input or, more recently, leveraging an attention mechanism \cite{bahdanau2014neural} that focuses only on parts of the input as well. This leaves the rest of the labeled examples unused for the current prediction, either during training or inference. 

In this work, we leverage similar examples in the training set to improve the performance and interpretability of deep neural networks, both at training and testing time. We propose an approach that retrieves the nearest training example to the one being processed and uses the corresponding target example (i) as auxiliary context to the input (e.g. combining the input together with the retrieved target), or (ii) to guide the attention mechanism of the neural network.

We show that the retrieved target can be easily incorporated in an LSTM model \cite{hochreiter1997long}, making use of its initial memory state. In general, previous studies have given little consideration to the initialization of the LSTM's memory state. Typically, the memory state is initialized simply with a vector of zeros. Even when it is initialized with the current context (e.g., the input image for captioning tasks), it is just initialized in the same way as the LSTM hidden state: with a simple affine transformation of the same context. Our approach takes advantage of the initial memory state by encoding auxiliary information from training examples.
We also present a new multi-level attention method that attends to the inputs and to the target of the nearest example.

We evaluate the proposed approach on 
image captioning and sentiment analysis. In brief, 
image captioning involves generating a textual description of an image. The dominant framework involves using a CNN as an encoder to represent the image, and passes this representation to a RNN decoder that generates the respective caption, combined with neural-attention \cite{xu2015show, hossain2019comprehensive}. In turn, sentiment analysis aims to classify the sentiment of an input text. Within neural methods, RNNs and CNNs are commonly used for sentiment analysis, recently also combining attention mechanisms \cite{poria2020beneath, ain2017sentiment}. 


Our general aim is to demonstrate the effectiveness of the proposed approach, which can also be easily integrated in other neural models, by applying it to standard methods for two different tasks (i.e., generation and classification) and by using a retrieval mechanism with different modalities (i.e., image and text retrieval).

\section{Related Work}
Our approach is closely related to those from some previous studies, in which models predict the output conditioned on retrieved examples \cite{hashimoto2018retrieve}\cite{weston2018retrieve}. For instance, Hashimoto et al. \cite{hashimoto2018retrieve} used an encoder-decoder model augmented by a learned retriever to generate source code. The authors suggested a retrieve-and-edit approach, in which the retriever finds the nearest input example and then that prototype is edited into an output pertinent to the input. In turn, Weston et al. \cite{weston2018retrieve} introduced a retriever for dialogue generation. The idea is to first retrieve the nearest response and then the generator, i.e. a seq-to-seq model, receives the current input concatenated with the retrieved response, separated with a special token. Our approach is similar, but rather than concatenating the input with the retrieved example, we make use of an LSTM’s memory cell state to incorporate the nearest training example. In our view, the retrieved examples should be considered as additional context, instead of being treated as regular input. Also, unlike Hashimoto et al. \cite{hashimoto2018retrieve}, 
our retrieval component does not need to be trained. 

Our retriever is, in fact, based on that from Khandelwal et al. \cite{khandelwal2019generalization}. In their work, a pre-trained Language Model (LM) is augmented with a nearest neighbor retrieval mechanism, in which the similar examples are discovered using FAISS \cite{JDH17}. 
The probabilities for the next word are computed by interpolating the LM's output distribution with the nearest neighbor distribution. However, differently from their work, we 
do not use the retriever just for inference, aiding the prediction both during training and testing time.

Retrieved examples have also been used to guide attention mechanisms. For instance Gu et al. \cite{gu2018search} described an attention-based neural machine translation model 
that attends to the current source sentence as well to retrieved translation pairs. There are also other multi-level attention studies that attend to more than the input. For instance, Lu et al. \cite{lu2017knowing} proposed an adaptive encoder-decoder framework for image captioning that can choose when to rely on the image or the language model to generate the next word. In turn, Li et al. \cite{li2020multi} proposed three attention structures, representing the attention to different image regions, to different words, and to vision and semantics. Our multi-level attention mechanism that leverages retrieval takes inspiration from these approaches. However, our attention mechanism is simpler and can be easily integrated on top of other attention modules. Our multi-level mechanism attends to the retrieved vector and to a given attention vector, which does not need to be the specific visual context vector that we use for image captioning or the context vector from the hidden states that we use for sentiment analysis. We can plug-in any model-specific attention vector to be attended together with the retrieved target vector. 

\section{Proposed Approach}

The proposed approach consists of two steps. The first step involves retrieving the nearest training example given the current input. The second step leverages the target of the nearest example, either by encoding it as the initial memory state of an LSTM or to guide an attention mechanism.

\subsection{Retrieval Component}

For retrieval, we use Facebook AI Similarity Search (FAISS) \cite{JDH17} to store the training examples as high-dimensional vectors, and search over the dataset. FAISS is an open source library for nearest-neighbor search, optimized for memory usage and speed, being able to quickly find similar high-dimensional vectors over a large pool of example vectors. To search over the stored vectors, FAISS uses as default the Euclidean distance, although it also supports the inner product distance. We use the default Euclidean distance to retrieve the nearest example $x_{n}$ given the current input $x_{c}$. 

Note that, for the second stage, the target output of the nearest example is required, and not the nearest example itself. This can be retrieved with an auxiliary lookup table that maps the corresponding index of $x_{n}$ to its target $y_{n}$. 

\paragraph{Image Retrieval} In image captioning, each training input image $x$ is stored as $D$-dimensional vector using a pre-trained encoder network:

\begin{equation}\label{eq1}
{r_{x}} = \mathrm{Enc(x)} \in \mathbb{R}^{D}.
\end{equation}

In the previous expression, ${r_{x}}$ denotes the vector representation of the input, and $D$ its dimensionality. A CNN encoder $Enc$ extracts the image features $V$ via the last convolutional layer, followed by a global average pooling operation. 

\paragraph{Text Retrieval} In sentiment analysis, each training input sentence $x$ is mapped to a vector representation ${r_{x}}$ using a pre-trained sentence representation model, denoted as S in the following expression: 

\begin{equation}\label{eq2}
{r_{x}} = \mathrm{S(x)}  \in \mathbb{R}^{D}.
\end{equation}

In particular, we use a pre-trained sentence transformer to obtain the corresponding sentence representations \cite{reimers2019sentence}, namely the \texttt{paraphrase-distilroberta-base-v1}\footnote{\label{roberta}\url{http://github.com/UKPLab/sentence-transformers}} model. This RoBERTa-based sentence representation model has been trained to produce meaningful sentence embeddings for similarity assessment and retrieval tasks. 

\subsection{Incorporating the Nearest Target in the LSTM}

In the second step, we use the target of the nearest example to initialize the memory state of an LSTM, or to guide an attention mechanism.

\subsubsection{LSTM Initial Memory State}
\label{memory-state}

\begin{figure*}[h!]
\begin{center}
  \includegraphics[width=0.96\linewidth]{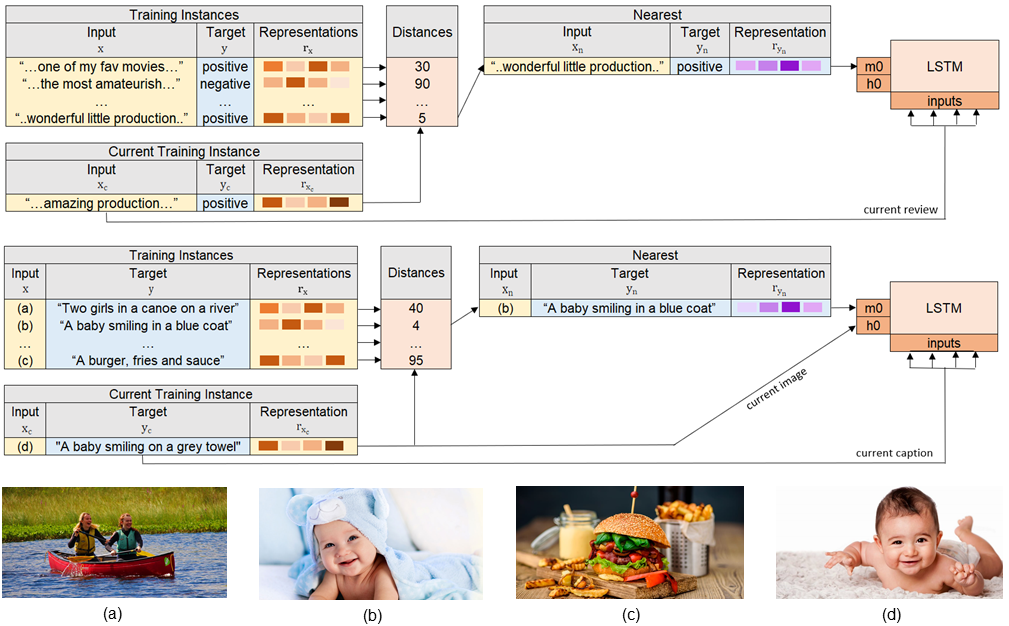}
 \end{center}
  \caption{Overview of the proposed approach. Typically, an LSTM model predicts the output based solely on the input, without leveraging auxiliar context from other training examples. Our approach retrieves the nearest training example $x_{n}$ and incorporates its target $y_{n}$ into the LSTM’s initial memory state $m_{0}$. We apply our approach to sentiment analysis (upper) and image captioning (down).}
  \label{fig:pic}
\end{figure*}

After retrieving the nearest input $x_{n}$ in the first step, the corresponding target $y_{n}$ is incorporated in the LSTM as the initial memory state (see Figure \ref{fig:pic}). This can be accomplished as long as the target example is encoded into a continuous vector space with the same dimensionality of the LSTM states. Our approach consists in mapping the retrieved target $y_{n}$ into a fixed-length representation and using an affine transformation to have the same dimensionality of the LSTM states. 

\begin{equation}\label{eq3}
r_{y_{n}}= W_{n}f(y_{n}).
\end{equation}

In the previous expression, $W_{n}$ is a learned parameter that projects the vector representation of the retrieved target to the same dimensionality of the LSTM. For the image captioning and sentiment analysis tasks, we use the vector representations $f(y_{n})$ produced through the procedures described next:

\paragraph{Image Captioning} We explore three alternative representations for the target caption of the nearest input image.
\begin{itemize}
    \item \textbf{Average of static word embeddings}: Each word is represented with a pre-trained embedding, in particular using fastText \cite{mikolov2018advances}, and then the word vectors are averaged to build the caption representation.
    
    \item \textbf{Weighted average of static word embeddings}:  Taking the average of word vectors assumes that all words are of equal importance. To better capture the meaning of a caption, we average the word embeddings weighted by their norms. Liu et al. \cite{liu2020norm} showed that a word embedding norm captures the word importance, with frequent words having smaller norms than rare words.
    
    \item \textbf{Contextual embeddings}: The aforementioned representations ignore word order and use static word embeddings not dependent on their left/right context within the sequence of words. To take this information into consideration, we encode the caption using the pre-trained sentence transformer \texttt{paraphrase-distilroberta-base-v1}\footref{roberta}.

\end{itemize}

\paragraph{Sentiment Analysis} In this task, the target of the nearest input sentence is either positive or negative. We suggest the following representations to encode the nearest target:
\begin{itemize}
\item \textbf{1s and -1s}: The nearest target is represented with a vector of 1s when positive, or with a vector of -1s when negative. The rationale behind this choice relates to using opposite vectors with a cosine similarity of -1. 
\item \textbf{Average embeddings of positive/negative sentences}: When the nearest target is positive, we use a representation obtained from all the positive training sentences through pre-trained fastText embeddings. This is done by averaging each word vector of each positive sentence and then averaging over all the positive sentences to obtain the final embedding that represents a positive target. The same idea is applied to a negative target (i.e., building a representation from the average embeddings of all negatives sentences). Essentially, we intend to provide the model with an overall memory of what resembles a positive/negative sentence. 
\item \textbf{Weighted average embeddings}: Similar to the previous formulation, but weighting each word by the norm of the corresponding embedding.
\item \textbf{Contextual embeddings}: Similar to the two aforementioned representations, but using \texttt{paraphrase-distilroberta-base-v1}\footref{roberta} to represent each sentence.
\end{itemize}

\subsubsection{Guiding an Attention Mechanism}
\label{multiattention}
The target of the nearest input can also be included in an attention mechanism. We present a new multi-level attention method that attends to the inputs and also to the retrieved target, deciding which of those to focus on for the current prediction.

\paragraph{Image Captioning} 

First, a visual context vector $c_{t}$ is computed with a typical additive attention \cite{bahdanau2014neural} given to the image features $V \in \mathbb{R}^{D \times K}$ and the previous hidden state $h_{t-1} \in \mathbb{R}^{D}$ of the LSTM, where $D$ and $K$ are respectively the dimensionality and the number of the image features:

\begin{equation}\label{eq4}
a_{t} = w_{a}^{T}\mathrm{tanh}(W_{v}V +W_{h}h_{t-1}),
\end{equation}
\begin{equation}\label{eq5}
\alpha_{t} =\mathrm{ softmax}(a_{t}),
\end{equation}
\begin{equation}\label{eq6}
c_{t} = \sum_{i=1}^{K} \alpha_{i,t} v_{i}.
\end{equation}

In the previous expressions, $W_{h}$, $W_{v}$, $w_{a}^{T}$ are learned parameters, and $\alpha_{t}$ are the attention weights. The attended image vector is defined as $c_{t}$, i.e. the visual context vector.

Then, the multi-level context vector $\hat{c_{t}}$ is obtained given the previous hidden state $h_{t-1}$ and the concatenation of the visual context vector  $c_{t}$ with the retrieved target vector $r_{y_{n}} \in \mathbb{R}^{D}$.

\begin{equation}\label{eq7}
\hat{a_{t}} = w_{\hat{a}}^{T}\mathrm{tanh} \left (W_{m} \mathrm{concat}(c_{t},r_{y_{n}}) + W_{h}h_{t-1} \right),
\end{equation}
\begin{equation}\label{eq8}
\hat{\alpha}_{t} = \mathrm{softmax}(\hat{a_{t}}),
\end{equation}
\begin{equation}\label{eq9}
\hat{c}_{t} = \hat{\alpha}_{1,t}c_{t} + \hat{\alpha}_{2,t}r_{y_{n}}.
\end{equation}

In the previous expressions, $W_{m}$ and $w_{\hat{a}}^{T}$ are parameters to be learned, while $\hat{\alpha}_{t}$ consists in $\hat{\alpha}_{1,t}$, i.e. the weight given to the visual context vector,  and $\hat{\alpha}_{2,t}$ is the weight  given to the retrieved target, with $\hat{\alpha}_{2,t}$=1- $\hat{\alpha}_{1,t}$. In this way, the proposed attention can decide to attend to the current image, or to focus on the nearest example.

\paragraph{Sentiment Analysis} First, a textual context vector is computed, similar to the attention vector described for image captioning. In this case, attention is not calculated at each time-step of the LSTM, since the prediction only occurs at final time-step $T$. Therefore, in Eq. \ref{eq4}, the previous hidden state $h_{t-1}$ is replaced by the last hidden-state $h_{t}$  and the visual features $V$ are replaced by all the LSTM hidden-states $H$.

\begin{equation}\label{eqlast}
a_{t} = w_{a}^{T}\mathrm{tanh}(W_{v}H +W_{h}h_{t}).
\end{equation}

Then, the proposed multi-level attention can decide to attend to the textual context vector (i.e. the alternative to the visual context vector for image captioning) or to focus on the retrieved target vector (Eq. \ref{eq7} to \ref{eq9}).

\section{Implementation Details}

We compare our two models, named Image Captioning through Retrieval (ICR) and Sentiment Analysis through Retrieval (SAR), against a vanilla encoder-decoder model with neural attention, and a vanilla attention-based LSTM, respectively. The differences between the baselines and our models are in the aforementioned approaches: (i) the LSTM’s memory state initialized with the target of the nearest input, and (ii) the multi-level attention mechanism. 

In image captioning, we use a ResNet model pre-trained on the ImageNet dataset as the encoder, without finetuning, together with a standard LSTM as the decoder, with one hidden layer and 512 units. For each input image, the ResNet encoder extracts the current image features $V=[v_{1},..., v_{k}]$ (2048 x 146$D$) and performs average pooling to obtain the global image feature $\bar{v}= \frac{1}{K}\sum_{i=1}^{K} v_{i}$ (that corresponds to $r_x$).  The initial hidden state of the LSTM is then initialized with an affine transformation of the image representation $h_0= W_{ih}\bar{v}$ (512$D$). The initial memory state $m_0$ is also initialized in the same way, with  $m_0= W_{im}\bar{v}$ for the baseline model. For the ICR model, $m_0$ is initialized with the retrieved target (Eq. \ref{eq3}), corresponding to the first reference caption of the retrieved nearest image. At each time-step $t$, the LSTM receives as input the fastText embedding of the current word in the caption (300$D$), concatenated with the corresponding attention context vector (512$D$). The baseline model uses the visual context vector $c_{t}$, and the ICR model uses the multi-level context vector $\hat{c_{t}}$. In the particular case of the multi-level attention mechanism, the image features $V$ receive an affine transformation before passing through Eq. \ref{eq4} to ensure the same dimensionality of the retrieved target in order to compute Eq. \ref{eq9}. Finally, the current output word probability is computed from the LSTM hidden state, processed through a dropout operation with rate of 0.5, and followed by an affine transformation and a final softmax normalization.

In sentiment analysis, we also use an LSTM with one hidden layer and 512 units. We use a standard LSTM for coherence with image captioning, but note that the SAR model can also use a bi-directional model. As typically done in sentiment analysis, the baseline LSTM hidden states are initialized with a vector of zeros, whereas our SAR model initializes $m_{0}$ with the sentiment of the nearest review (Eq. \ref{eq3}).  
At each time step, the LSTM receives as input the current word embedding (based on pre-trained fastText embeddings). After processing the whole sequence, at the last time-step $T$, attention is applied over the respective hidden states $H$ and the last hidden state $h_t$. Then, the corresponding context vector is processed through a dropout operation with rate of 0.5, followed by an affine transformation and a sigmoid activation, in order to obtain the probability for the positive class.

The models are trained with the standard categorical and binary cross-entropy loss, respectively for image captioning and sentiment analysis. The batch size is set to 32 and we use the Adam optimizer with a learning rate of 4e-4 and 1e-3, respectively for image captioning and sentiment analysis. As stopping criteria, we use early stopping to terminate if there is no improvement after 12 consecutive epochs on the validation set (over BLEU score for image captioning, and accuracy for sentiment analysis) and the learning rate is decayed after 5 consecutive epochs without improvement (shrink factor of 0.8). At testing time, we use greedy decoding in image captioning. All our experiments were executed on a workstation with a single NVIDIA TITAN Xp GPU, with an Intel i7 CPU and 64Gb of RAM.

\section{Evaluation}
This section presents the experimental evaluation of the proposed approach. We first describe the datasets and evaluation metrics, and then present and discuss the obtained results. 

\subsection{Datasets and Metrics}

We report experimental results on commonly used datasets for image captioning and sentiment analysis, namely the Flickr8k dataset \cite{hodosh2013framing} and the IMDB dataset \cite{maas2011learning}, respectively. Flickr8k has 8000 images with 5 reference captions per image. We use the publicly available splits of Karpathy\footnote{\url{http://cs.stanford.edu/people/karpathy/deepimagesent/}}, with 6000 images for training, 1000 images for validation, and the remaining 1000 images for testing. The IMDB dataset contain 50000 movie reviews from IMDB, labeled as positive or negative. We use the publicly available train-test splits, with 25000 reviews each, and we do an additional random split on the training set with 10\% of the original training reviews for validation. For both datasets, the vocabulary consists of words that occur at least five times.



To evaluate caption quality, we use common image captioning metrics in the literature such as BLEU, METEOR, ROUGE\_L, CIDEr and SPICE~\cite{celikyilmaz2020evaluation}. All the aforementioned metrics were calculated through the implementation in the MS COCO caption evaluation package\footnote{\url{http://github.com/tylin/coco-caption}}. 
Additionally, we use the recent BERTScore~\cite{zhang2019bertscore} metric, which has a better correlation with human judgments. 
Regarding the evaluation of sentiment analysis, we also use established metrics, namely the F-score and the classification accuracy. 

\subsection{Image Captioning Results}

\begin{table*}
  \centering
    \caption{Image Captioning performance on the Flickr8k test split, with the best results shown in bold.}

  \begin{tabular}{lccccccccc}
    \hline
    Models        & BLEU-1 & BLEU-2 & BLEU-3 & BLEU-4 &METEOR  & ROUGE\_L &  CIDEr & SPICE & BERTScore  \\
    \hline
    Baseline  &0.5541&0.3848&0.2554&0.1691&0.1948&0.4421&0.4526&0.1353 &0.4945\\
    ICR \emph{avg}  &0.6044&0.4199&0.2838&\textbf{0.1921}&0.1995&0.4537&0.4815&\textbf{0.1400}&0.5141      \\
    ICR \emph{weighted}  &\textbf{0.6080}&\textbf{0.4251}&\textbf{0.2857}&0.1896&\textbf{0.2002} &\textbf{0.4583}&\textbf{0.4897}&0.1364&\textbf{0.5250}      \\
    ICR \emph{RoBERTa}~~~~~~~~~~~~~~~~~~~~~~~~ &0.5831&0.4023&0.2689&0.1788&0.1992&0.4476&0.4643&0.1373&0.5064\\
    \hline
    \hspace*{-1cm}
  \end{tabular}

  \label{ICR1}

\end{table*}

\begin{table*}
  \centering
  \caption{Ablation study for Image Captioning. The baseline is compared against using the retrieved target in the LSTM's initial memory (\emph{initialization $m_{0}$}) and in the neural attention (\emph{multi-level attention}). Best results are shown in bold and the second-best underlined.}
  \begin{tabular}{lccccccccc}
    \hline
    Models        & BLEU-1 & BLEU-2 & BLEU-3 & BLEU-4 &METEOR  & ROUGE\_L &  CIDEr & SPICE& BERTScore  \\
    \hline
    Baseline  &0.5541&0.3848&0.2554&0.1691&0.1948&0.4421&0.4526&0.1353&0.4945 \\
    ICR \emph{weighted initialization $m_{0}$} &\textbf{0.5971}&\textbf{0.4188}&\textbf{0.2817}&\textbf{0.1877}&\underline{0.2016}&\textbf{0.4548}&\underline{0.4822}&\textbf{0.1377}&\textbf{0.5263}\\
    ICR \emph{weighted multi-level attention} & \underline{0.5908}&\underline{0.4134}&\underline{0.2782}&\underline{0.1873}&\textbf{0.2030}&\underline{0.4534}&\textbf{0.4926}&\underline{0.1371}&\underline{0.5260}\\
       \hline
      ICR \emph{weighted} (combined) &0.6080&0.4251&0.2857&0.1896&0.2002 &0.4583&0.4897&0.1364&0.5250      \\
    \hline
  \end{tabular}

 \label{ICR2}
\end{table*}

In Table \ref{ICR1}, we present the captioning results on the Flickr8k dataset. We compare the baseline with the proposed ICR model, displaying the performance of the different representations suggested to encode the retrieved target caption, namely using average pre-trained embeddings (\emph{avg}), weighted average (\emph{weighted}) and RoBERTa sentence embeddings (\emph{RoBERTa}). Results show that the ICR model outperforms the baseline, achieving a better performance on all metrics independently of the representation being used. The best performance was achieved using the nearest target encoded with the fastText weighted average embeddings, followed by simple average and then by RoBERTa embeddings, with all the three surpassing the baseline. We hypothesise that static fastText embeddings worked better than contextual RoBERTa embeddings in the representation of the retrieved target possibly due to the use of fastText embeddings also for the generated words, i.e. lying on the same representation space that is used by the decoder. 

\begin{figure}[!ht]
  \includegraphics[width=1.0\linewidth]{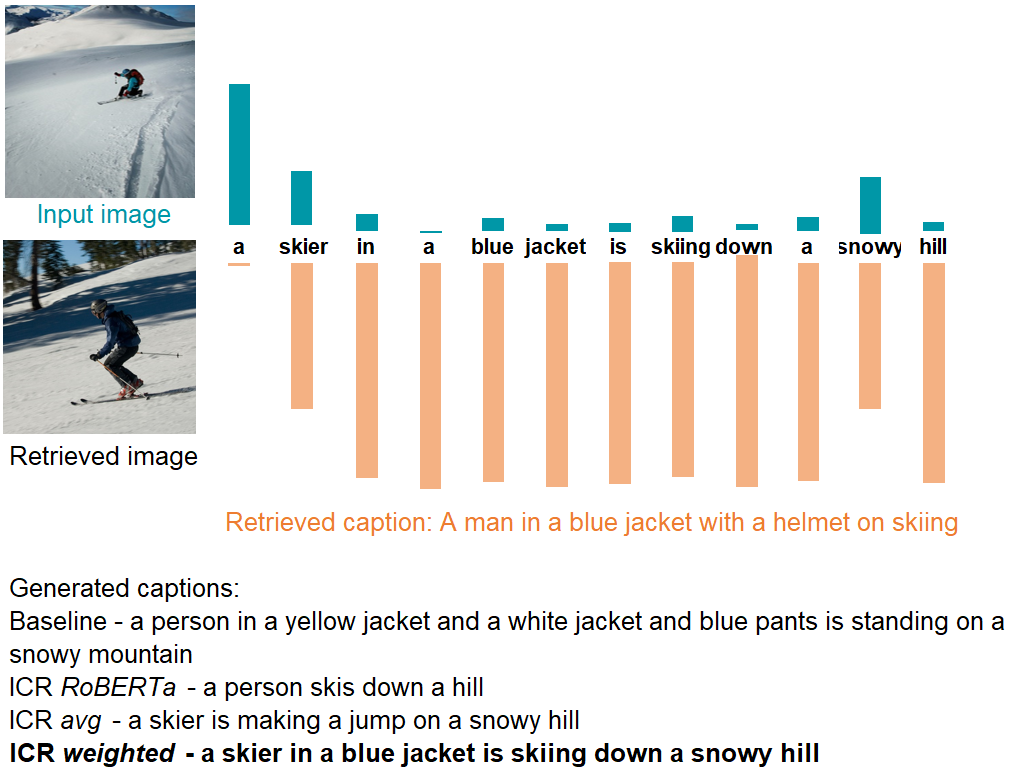}
  \centering
  \caption{Upper: A visualization of our multi-level attention mechanism, showing how much attention the ICR \emph{weighted} model pays to the current input image (blue bars) and to the retrieved target caption (orange bars). Bottom: the captions generated by the different models for the input image.}
  \label{fig:ic_example}
\end{figure}

\begin{figure}[!ht]
  \includegraphics[width=1.0\linewidth]{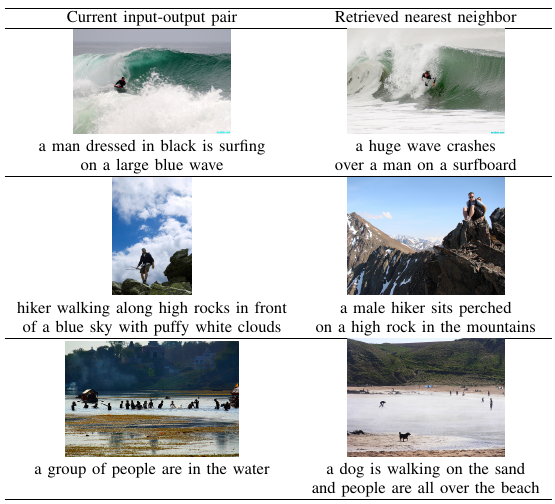}
  \centering
  \caption{Examples of retrieved images, together with the corresponding target captions used in our ICR model.}
  \label{fig:ic-appendix2}
\end{figure}

\begin{figure*}[!ht]
 \includegraphics[width=0.89\linewidth]{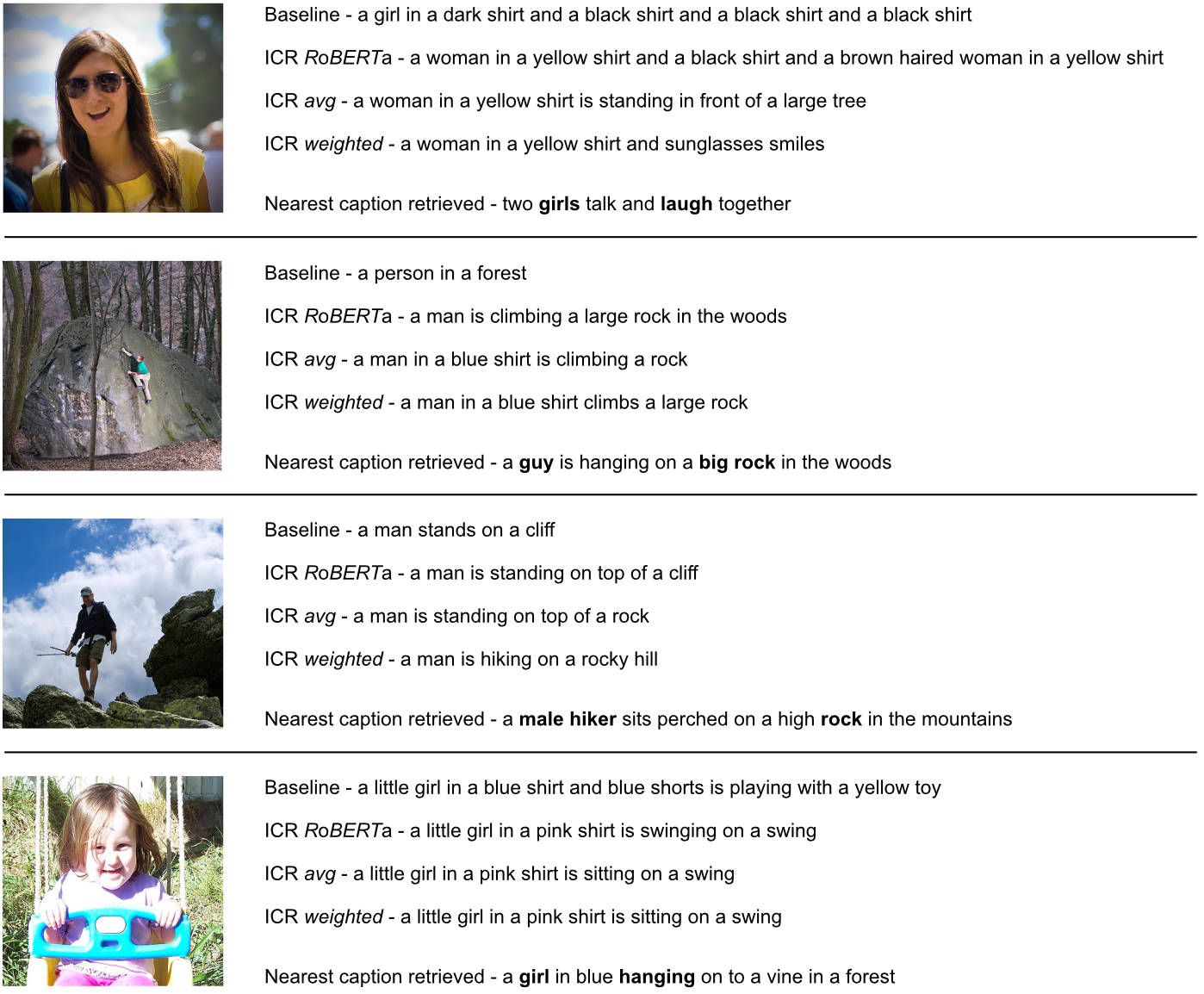}
  \caption{Examples of test captions generated by the different models under analysis. The captioning quality tends to improve using the semantics of the retrieved target captions, with relevant words highlighted in bold over the examples.}
  \label{fig:pic-new}
\end{figure*}

A representative example is shown in Figure \ref{fig:ic_example}, containing the captions generated by the aforementioned models, together with a visualization of our multi-level attention mechanism from the best model (ICR \emph{weighted}). We provide more examples in Figure \ref{fig:pic-new}. These qualitative results further confirm that the ICR \emph{weighted} model tends to produce better captions. Regarding the multi-level attention mechanism, in general, we noticed that the ICR \emph{weighted} model tends to attend to the current input image in the beginning of the generation process and, after obtaining the current context, the model switches its attention to focus more on the semantics of the nearest caption. Figure \ref{fig:pic-new} also shows that the captions generated by our model express the semantics of the retrieved target captions. As observed in these examples, the retrieved information can aid the prediction and it can also provide some evidence for the model's decision. We thus argue that augmenting LSTM models with retrieved examples can also bring benefits in terms of model interpretability (i.e., users can better understand the predictions by analyzing the retrieved examples that guided them). We also provide examples for the retrieved nearest instances in Figure \ref{fig:ic-appendix2}. We can see that, sometimes, the retrieved images are not that similar to the given image, or they have some similar aspects but with different caption contexts (see the last row). Better image representations could be achieved by fine-tuning the encoder on a task derived from the information associated to the captions (e.g. predicting the nouns and adjectives of the captions, thus promoting similarity between images that correspond to similar captions), instead of just relying on ImageNet pre-training.

We also performed ablation studies, as shown in Table \ref{ICR2}, to quantify the impact of each method we use in the ICR model: the retrieved target incorporated in the LSTM’s memory state (\emph{initialization $m_{0}$}) and the multi-level attention mechanism (\emph{multi-level attention}). Specifically we compare the baseline against the best performing model (ICR \emph{weighted}), either including the former or the later method. Observing Table \ref{ICR2}, the baseline obtains the lowest scores across all metrics. The performance is improved using the retrieved information in the LSTM memory cell or in the attention, showing that the retrieved targets are effectively exploited in both methods. 

Overall, in terms of quantitative and qualitative results, we observed performance gains by incorporating the nearest target in the LSTM initial memory state or in the neural attention mechanism used, compared with a baseline without the retrieved information. Both the ICR model and the baseline consisted of standard encoder-decoder models with neural attention, but the proposed approach could be also integrated in stronger methods, such as the widely used bottom-up and top-down attention model \cite{anderson2018bottom}.

\subsection{Sentiment Analysis Results}

Table \ref{SAR1} presents the sentiment analysis results on the IMDB dataset. We contrast the baseline model against the proposed SAR model, comparing the performance of the different representations suggested to encode the retrieved target label, consisting in using vectors of -1s and 1s (\emph{-11s}) to represent the negative and positive labels, respectively, or using the average of fastText embeddings (\emph{avg}) of all the positive and negative training reviews, their weighted average (\emph{weighted}), or RoBERTa sentence embeddings (\emph{RoBERTa}). While the performance decreases with the SAR \emph{-11s} model compared to the baseline, suggesting that this representation is not effective, the other SAR models slightly outperform the baseline, yielding a better performance on both accuracy and F-score. Our best result, using the SAR \emph{avg} model, yields a 0.053 increase in accuracy compared to the baseline and a 0.052 increase in the F-score. We also conducted ablation studies with the proposed methods (i.e., \emph{initialization $m_{0}$} or \emph{multi-level attention}), and observed that both outperformed the baseline, as can be seen in Table \ref{SAR2}. It should also be noticed that the differences in performance are very small, but perhaps with a more challenging dataset or in another classification task the influence could be larger. 

\begin{table}[!ht]
  \centering
   \caption{Sentiment Analysis results on the IMDB test split. Best performance is shown in bold.}
  \begin{tabular}{lcc}
    \hline
    Models        & Accuracy & F-score  \\
    \hline
    Baseline  &0.8930&0.8929\\
    SAR \emph{-11s} & 0.8914&0.8913\\
    SAR \emph{avg}  &\textbf{0.8983}&\textbf{0.8981}\\
    SAR \emph{weighted} &0.8978&0.8976\\
    SAR \emph{RoBERTa}~~~~~~~~~~~~~~~~~~~~~~~~~ &0.8963&0.8961\\
    
    \hline
  \end{tabular}
 
  \label{SAR1}
\end{table}

\begin{table}[!ht]
  \centering
  \caption{Ablation study for Sentiment Analysis. Best performance is shown in bold and the second-best is underlined.}
  \begin{tabular}{lcc}
    \hline
    Models        & Accuracy & F-score  \\
    \hline
    Baseline  &0.8930&0.8929\\
    SAR \emph{avg initialization $m_{0}$}&\underline{0.8935}&\underline{0.8933}\\
    SAR \emph{avg multi-level attention}~~~~~~~~~&\textbf{0.8949}&\textbf{0.8948} \\
    \hline
    SAR \emph{avg} (combined) &0.8983&0.8981\\
    
    \hline
  \end{tabular}
  
  \label{SAR2}
\end{table}

\label{sa-app}
\begin{table*} [!ht]
  \centering
  \caption{Examples of reviews with their corresponding labels, and the retrieved nearest reviews with their labels. (+) denotes a positive review and (-) denotes a negative review.}
  \begin{tabular}{l l}
    
    Current input-output pair        & Retrieved nearest neighbor  \\
    \hline

    (\textbf{+}) i have watched it  at least  twenty times and (...) \underline{absolutely perfect} (...) & (\textbf{+}) it s very funny it has a great cast who each give \underline{great performances} \\ 
     
      filled with stars who perform their roles to  perfection  \underline{kevin kline} (...) &especially sally field and \underline{kevin kline} it s a well written screenplay (...) \\

     \hline
    (\textbf{-})  i can t say much about this film i think it speaks for itself as do the   & (\textbf{-}) \underline{this movie was a disappointment} i was  looking forward to seeing a  \\ 
     
      current ratings on  here i rented this  about two years ago and \underline{i  totally}&  good movie am the type of person who starts a movie and doesn t turn \\
     
      \underline{regretted} it (...)  & it off until the end but  i was forcing myself not to turn it off (...)\\

     \hline
     (\textbf{+})  obviously written for the stage lightweight but \underline{worthwhile} how can    & (\textbf{+}) (...) \underline{olivier} s dramatic performances (...) in this film he is disarmed \\ 
     
    you go wrong with ralph richardson \underline{olivier and merle oberon}&  of his pomp and overconfidence by sassy \underline{merle oberon} and plays the   \\

     & flustered divorce attorney underline{with great charm}\\

      \hline
     (\textbf{-})  by the numbers \underline{story of the kid prince} a  singer on his way to becoming   & (\textbf{+})  purple rain has never been a critic s darling but it is a cult classic  \\ 
     
        a star then he falls in love with appolonia kotero but he has to deal with his  & and deserves to be \underline{if you are a prince fan this is for you} the main plot\\
     
     wife beating father (...) i couldn t believe it \underline{the script is terrible lousy}  &  is prince seeing his abusive parents in himself and him falling in love \\
     
        \underline{dialogue} and some truly painful comedy routines (...) & with a girl (...)\\

    \hline
  \end{tabular}
  
  \label{SAR3}
\end{table*}

Some retrieval examples are presented in Table \ref{SAR3}. In most cases, the text retrieval is effective in finding a good neighbor, but sometimes the neighbor can be the same movie with an opposite review, as we can see in the last example.


\section{Conclusions and Future Work}

In this work, we proposed a new approach that can be incorporated into any LSTM model, leveraging the information from similar examples in the training set, during training and test time. For a given input, we first retrieve the nearest training input example and then use the corresponding target as auxiliar context to the input of the neural network, or to guide an attention mechanism. We showed that the retrieved target can be easily incorporated in an LSTM initial memory state, and we also designed a multi-level attention mechanism that can attend both to the input and the target of the nearest example. We conducted a series of experiments, presenting alternative ways to represent the nearest examples for two different tasks, image captioning and sentiment analysis. In both cases, our approach achieved better results than baselines without the retrieved information. Besides aiding to improve result quality, our retrieval approach also provides cues for the model's decisions, and thus it can also be of help in terms of making models more interpretable.


Despite the interesting results, there are also many possible ideas for future work. For instance, further work could better explore our approach in terms of interpretability. Additionally, both models could be improved by receiving information from the top nearest examples, instead of relying on a single retrieved example. In the particular case of image captioning, the encoder could be fine-tuned to produce better image representations, this away also improving the search for nearest examples. 
In addition, the selection of the retrieved target
could also be improved. In this work, we simply select the first reference  of the retrieved input image, when there are actually five possible references for the retrieved target. Furthermore, the proposed multi-level attention mechanism could attend to the various words of the retrieved caption, as it does for the image regions. Regarding sentiment analysis, it would be interesting to conduct experiments in more challenging and fine-grained datasets than IMDB positive/negative reviews to better assess the proposed approach. We also note that, in movie reviews, neighbor inputs can have different targets while nonetheless being similar, since they refer to the same movie. It would be interesting to explore the proposed approach in others classification tasks, where neighbor inputs tend to have more similar targets.

\section*{Acknowledgements}

We would like to thank André F. T. Martins and Marcos Treviso for the feedback on preliminary versions of this document. Our work was supported by national funds through Fundação para a Ciência e a Tecnologia (FCT), under grants with references UIDB/50021/2020 (INESC-ID multi-annual funding), PTDC/CCI-CIF/32607/2017 (MIMU), POCI/01/0145/FEDER/031460 (DARGMINTS), and the Ph.D. scholarship with reference 2020.06106.BD; through Agência Nacional de Inovação (ANI), under the project CMU-PT Ref. 045909 (MAIA); and through European Union funds (Multi3Generation COST Action CA18231). We also acknowledge the support of NVIDIA Corporation, with the donation of the TITAN Xp GPU used in our experiments.



\end{document}